\documentclass{article}

\usepackage{PRIMEarxiv}

\usepackage[utf8]{inputenc} 
\usepackage[T1]{fontenc}    
\usepackage{hyperref}       
\usepackage{url}            
\usepackage{booktabs}       
\usepackage{amsfonts}       
\usepackage{nicefrac}       
\usepackage{microtype}      
\usepackage{lipsum}
\usepackage{fancyhdr}       
\usepackage{graphicx}       
\graphicspath{{media/}}     

\usepackage{amsmath}
\usepackage{bm}
\usepackage{algorithm}
\usepackage{algpseudocode}
\newcommand{\vtheta}{v_{\!\theta}}
\newcommand{\vtilde}{\tilde{\bm{v}}}
\newcommand{\norm}[1]{\left\|#1\right\|}
\newcommand{\E}{\mathbb{E}}

\title{Velocity Scheduled Flow Matching
}

\author{
  Vitalii Bondar \\
  Cherkasy State Technological University \\
  Cherkasy, Ukraine \\
  v.bondar@chdtu.edu.ua \\
}

\begin{document}
\maketitle

\begin{abstract}
Flow matching trains a neural network to regress the conditional velocity
along a linear interpolant between noise and data, and the number of network
evaluations~(NFE) sets the cost of sampling. The straight-line interpolant
carries an implicit choice: the sample moves at constant speed throughout
the trajectory. We relax this choice and introduce Velocity Scheduled Flow
Matching~(VSFM), which replaces the conditional target $x_1 - x_0$ with
$v(t)(x_1 - x_0)$ for any nonnegative profile
$v:[0,1]\to\mathbb{R}_{\geq 0}$ satisfying $\int_0^1 v\,dt = 1$. We study
six polynomial profiles drawn from motion planning. The first use of VSFM
is at inference time: a pretrained linear flow-matching model can be
sampled under any admissible profile by integrating its ODE on a
non-uniform $\tau$-schedule, with no retraining and no additional
computation; on CIFAR-10 this lowers FID by up to $19.8\%$. Training from
scratch under a braking profile gives a further reduction of $17.4\%$ at
$4$~NFE. Both gains follow from the local truncation error of the Euler
integrator on the induced grid.
\end{abstract}


\section{Introduction}
\label{sec:introduction}

Flow matching is a continuous-time generative modeling framework 
with applications in image synthesis~\cite{lipman2022flow,liu2022flow,tong2024improving}
and scientific simulation~\cite{klein2023equivariant,bose2024foldflow}.
A neural velocity field $\vtheta(x,t)$ is trained to match the conditional velocity
along an interpolant between noise $x_0\sim p_0$ and data $x_1\sim p_1$,
and samples are drawn by integrating the learned ODE $dx/dt=\vtheta(x,t)$,
so that the trained model realizes a learned transformation from the noise distribution to the data distribution~\cite{bondar2025deep}.
The sampling cost is set by the number of network evaluations (NFE)
needed to integrate the ODE accurately.
This NFE budget is the inference-time compute that can be traded for sample quality, an instance of inference-time scaling~\cite{bondar2026inference}.
The low-NFE regime, in which a handful of Euler steps must carry the trajectory from noise to data, 
is the active frontier for fast samplers built on pretrained models, 
including large-scale text-to-image systems~\cite{rombach2022high,esser2024scaling}.

Existing flow-matching models fix the interpolant to the straight line $x_t = x_0 + t(x_1 - x_0)$. 
The choice is convenient: the conditional velocity reduces to the time-independent displacement $x_1 - x_0$,
probability-flow trajectories are straight and admit an optimal-transport interpretation~\cite{lipman2022flow,shaul2023kinetic}.
It also commits the sampler to moving at \emph{constant speed} from noise to data, 
a property that prior work has not isolated as a design choice.

We generalize the conditional target from a constant to a time-varying form,
\begin{equation}
u_t(x \mid x_0, x_1) = v(t)\cdot(x_1 - x_0),
\label{eq:vsfm-target-intro}
\end{equation}
where $v:[0,1]\to\mathbb{R}_{\geq 0}$ is a velocity profile satisfying $\int_0^1 v(t)\,dt = 1$. 
The unit integral preserves total displacement, so every admissible $v$ transports noise to data; 
the remaining freedom is the rate at which the sample moves at each instant. 
Setting $v(t)\equiv 1$ recovers standard flow matching. 
We call the resulting framework \emph{Velocity Scheduled Flow Matching} (VSFM); 
the architecture, loss, coupling, and integrator are otherwise unchanged.
Section~\ref{sec:method-profiles} instantiates VSFM with six polynomial profiles drawn 
from boundary-value problems in motion planning.

Two consequences follow from this generalization. The first is a training-free inference-time rescheduling: 
any admissible profile can be applied to a pretrained linear flow-matching model 
by running its ODE on the non-uniform step schedule that the profile induces (Section~\ref{sec:method-inference}). 
The second is training from scratch under a non-constant profile, 
which amounts to a one-line modification of the standard loss and yields additional FID gains, 
with the largest effect at low NFE.

On CIFAR-10, training from scratch under the coasting profile lowers FID by 17.4\% at 4~NFE and 10.4\% at 50~NFE relative to the linear baseline.
Applied as an inference-time modification to the publicly available TorchCFM checkpoint~\cite{tong2024improving}, 
the same profile lowers FID by up to 19.8\% across the NFE budgets we evaluated, without retraining. 
Section~\ref{sec:method-error} traces both effects to the local truncation error of the Euler sampler in the $\tau$-space ODE.

The paper is organized as follows. Section~\ref{sec:related-work} positions VSFM against prior work. 
Section~\ref{sec:method} develops the training target, derives the inference-time rescheduling algorithm, 
and analyses the discretization error it induces. 
Section~\ref{sec:experiments} reports the CIFAR-10 experiments, 
and Section~\ref{sec:conclusions} concludes and outlines directions for follow-up work.

\section{Related Work}
\label{sec:related-work}

VSFM connects to four threads of prior work: the choice of interpolant in flow matching, 
kinetic-energy analyses of probability paths, noise schedules and loss weighting in diffusion, 
and time-sampling strategies at training and inference.

\subsection{Flow matching and the linear interpolant}

Flow matching was introduced by Lipman et al.~\cite{lipman2022flow} and Liu et al.~\cite{liu2022flow} 
and extended to arbitrary couplings by Tong et al.~\cite{tong2024improving}. 
In all three formulations the conditional interpolant is the straight line $x_t=x_0+t(x_1-x_0)$, 
a choice retained by most subsequent work, including the rectified-flow transformer underlying Stable Diffusion~3~\cite{esser2024scaling}. 
Rectified flow~\cite{liu2022flow} straightens the learned flow by iterative re-coupling; 
Consistency Flow Matching~\cite{yang2024consistencyfm} enforces a self-consistency constraint on the velocity field; 
both stay within the linear interpolant. 
SplineFlow~\cite{rathod2026splineflow} replaces the linear interpolant 
with a B-spline construction in a multi-marginal dynamical-systems setting, 
targeting different goals and offering no inference-time adaptation for pretrained models. 
Albergo and Vanden-Eijnden~\cite{albergo2022building} and Pooladian et al.~\cite{pooladian2023multisample} 
study stochastic interpolants and multisample flow matching at the level of theory; 
both modify the coupling and the probability path, 
while VSFM alters the speed at which the trajectory is traversed. 
VSFM also changes the interpolant but does so for two-marginal generative modeling, 
uses a simpler polynomial family, and admits a training-free inference variant.

\subsection{Kinetic-energy perspectives on flow matching}

Shaul et al.~\cite{shaul2023kinetic} identify the Gaussian probability path 
that minimizes kinetic energy $\int_0^1\E[\|u_t\|^2]\,dt$ within a class of affine conditionals, 
and show that the straight-line constant-velocity path is approximately optimal under that criterion.
VSFM optimizes a different quantity, namely the local truncation error of the Euler integrator at a fixed step budget. 
That criterion depends on how discretization error is distributed along $t$, 
rather than on integrated squared velocity, 
and Section~\ref{sec:method-error} demonstrates that a non-constant profile is preferred. 
The two results are compatible: constant velocity minimizes kinetic energy, 
while a non-constant velocity lowers Euler error at fixed NFE. 
Neither the kinetic-energy formulation nor the rectified-flow literature provides a
way to adapt the sampling dynamics of an already-trained model;
Section~\ref{sec:method-inference} supplies one.

\subsection{Noise schedules and loss weighting in diffusion}

The design of the forward process in diffusion models has long been recognized as influential on sample quality. 
Sohl-Dickstein et al.~\cite{pmlr-v37-sohl-dickstein15} introduced the diffusion framework; 
Ho et al.~\cite{ho2020denoising} scaled it with a tractable denoising objective and a linear variance schedule. 
Song et al.~\cite{song2021scorebased} unified diffusion and score-based models as SDEs, 
placing the ODE-based inference used by flow matching in a broader probabilistic setting. 
Dhariwal and Nichol~\cite{dhariwal2021diffusion} showed that classifier guidance enables diffusion models to reach high
sample quality on class-conditional benchmarks. 
Nichol and Dhariwal~\cite{nichol2021improved} proposed a cosine schedule 
that avoids degeneracy near the pure-noise endpoint. 
Karras et al.~\cite{karras2022elucidating} separated and independently tuned the
forward process, the sampler, and the training components. 
P2 weighting~\cite{choi2022perception} and Min-SNR weighting~\cite{hang2023efficient} reweight 
the per-noise-level training loss, 
and Kingma and Gao~\cite{kingma2023understanding} give a unifying likelihood interpretation of such weighted losses. 
VSFM modifies the interpolant rather than the schedule or a loss coefficient, 
and, unlike loss-weighting methods, admits an inference-time variant that requires no retraining.

\subsection{Time sampling and step allocation}

A separate line of work alters which times $t$ are sampled during training, 
or the discretization grid at inference, while leaving the linear interpolant fixed. 
Esser et al.~\cite{esser2024scaling} bias training-time sampling toward the center of $[0,1]$ 
through a logit-normal distribution. 
Karras et al.~\cite{karras2022elucidating} design non-uniform inference-time grids for diffusion ODEs. 
Dense-Jump Flow Matching~\cite{chen2025densejump} combines U-shaped training-time
sampling with a jump-based solver for robotic policies. 
All three perturb $p(t)$ or the discretization grid while keeping the interpolant linear.

The inference procedure of Section~\ref{sec:method-inference} reduces,
in mechanical terms, to running the pretrained sampler on the non-uniform schedule $\tau_i = \alpha(i/N)$, 
and therefore belongs to this family.
What VSFM adds is a structured parameterization of the schedule through the velocity profile $v(t)$, 
a training-time variant that aligns the model with its inference schedule, 
and a discretization-error account (Section~\ref{sec:method-error}) 
that singles out the braking sub-family without a grid search.

\subsection{Polynomial movement profiles in motion planning}

Polynomial movement profiles are standard tools in robotics, animation, 
and CNC machining for generating smooth actuator paths under boundary conditions on velocity and acceleration~\cite{craig2009introduction}. 
Flash and Hogan~\cite{flash1985coordination} showed that 
the minimum-jerk profile uniquely minimizes integrated squared jerk subject to zero velocity and acceleration at both endpoints, 
and it has remained a standard model of unconstrained human reaching movements ever since.
These kinematic profiles have not previously been connected to generative modeling.

\section{Method}
\label{sec:method}

\subsection{Generalized training target}
\label{sec:method-generalized}

Let $v:[0,1]\to\mathbb{R}_{\geq 0}$ be a nonnegative function satisfying
\begin{equation}
  \int_0^1 v(t)\,dt = 1.
  \label{eq:v-integral}
\end{equation}
We call any such $v$ an \emph{admissible velocity profile}. 
Nonnegativity rules out backtracking, and the unit-integral condition guarantees that
the resulting motion transports $x_0$ to $x_1$. 
The generalized conditional velocity field is
\begin{equation}
  u_t(x \mid x_0, x_1) = v(t)\cdot(x_1 - x_0),
  \label{eq:vsfm-conditional}
\end{equation}
the form introduced in Section~\ref{sec:introduction}; 
the constant target of standard flow matching is recovered at $v\equiv 1$. 
The associated \emph{movement profile} $\alpha(t) = \int_0^t v(s)\,ds$ satisfies
$\alpha(0)=0$, $\alpha(1)=1$, and $\alpha'(t)=v(t)$, 
and the sample location at time $t$ is the affine interpolant
\begin{equation}
  x_t = x_0 + \alpha(t)\,(x_1 - x_0).
  \label{eq:vsfm-interpolant}
\end{equation}
The displacement $x_1 - x_0$ is fixed by the coupling; 
$\alpha(t)$ records the fraction of that displacement already traversed; 
and $v(t)$ controls the rate at which it accumulates. 
The VSFM training objective regresses the network on the displacement $x_1-x_0$ 
at the profile-dependent positions $x_t$:
\begin{equation}
  \mathcal{L}_{\mathrm{VSFM}}(\theta)
  = \E_{t,x_0,x_1}\!\left[\norm{\vtheta(x_t, t) - (x_1 - x_0)}^2\right],
  \label{eq:vsfm-loss}
\end{equation}
with $t \sim \mathcal{U}[0,1]$ and $(x_0,x_1)$ drawn from the coupling.
Setting $v\equiv 1$ recovers the standard flow-matching loss. 
At inference, the conditional velocity field is reconstructed as
$v(t)\,\vtheta(x_t,t)$ (Section~\ref{sec:method-inference}).

Equation~\eqref{eq:vsfm-loss} is equivalent to the natural form
$\E[\norm{\vtheta - v(t)(x_1-x_0)}^2]$ with the explicit $v(t)^2$ weighting absorbed. 
We adopt the unweighted form because regressing on the constant-magnitude displacement 
is numerically better behaved under braking profiles, 
where the velocity target $v(t)(x_1-x_0)$ is amplified near $t=0$ and collapses to zero near $t=1$. 
In our runs the $v(t)^2$-weighted loss destabilized gradient updates over the course of training.

The velocity profile then enters training through the distribution of positions $x_t$ 
rather than the magnitude of the regression target. 
With $t\sim\mathcal{U}[0,1]$, the fraction $\alpha(t)$ is itself non-uniformly
distributed and concentrates training examples where the profile moves fastest. 
For braking profiles this puts more training mass on positions close to the data endpoint, 
where the network refines fine-grained structure. 
Algorithm~\ref{alg:vsfm-training} states the resulting training loop in full.

\begin{algorithm}[h]
\caption{VSFM training.}
\label{alg:vsfm-training}
\begin{algorithmic}[1]
\Require Network $\vtheta$; movement profile $\alpha$; training steps $K$
\For{step $k = 1, 2, \ldots, K$}
  \State Sample $t\sim\mathcal{U}[0,1]$, $(x_0,x_1)\sim p(x_0,x_1)$
  \State $x_t \gets x_0 + \alpha(t)\,(x_1-x_0)$
        \Comment{interpolant via movement profile}
  \State $\ell \gets \norm{\vtheta(x_t,\,t) - (x_1-x_0)}^2$
        \Comment{Equation~\eqref{eq:vsfm-loss}}
  \State Update $\theta$ by gradient descent on $\ell$
\EndFor
\end{algorithmic}
\end{algorithm}

\subsection{Polynomial velocity profiles}
\label{sec:method-profiles}

\begin{figure}
  \centering
  \includegraphics{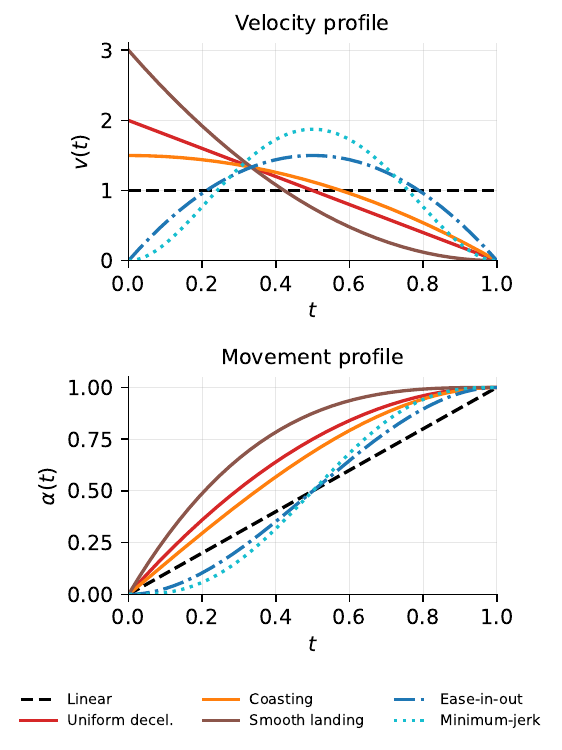}
  \caption{Velocity profiles $v(t)$ (top) and the corresponding movement profiles $\alpha(t)=\int_0^t v(s)\,ds$ (bottom) for the six profiles of Table~\ref{tab:profiles}.}
  \label{fig:profiles}
\end{figure}

We instantiate VSFM with six polynomial velocity profiles: the linear baseline,
three braking profiles (uniform deceleration, coasting, and smooth landing),
and two symmetric profiles (ease-in-out and minimum-jerk).
Each one is obtained by imposing boundary conditions on $\alpha$
and its derivatives at the endpoints of $[0,1]$, 
conditions that pin down a polynomial of minimal degree,
in line with standard practice in motion planning~\cite{flash1985coordination,craig2009introduction}.
Requiring more derivatives of $\alpha$ to vanish at an endpoint raises the smoothness of the profile there,
at the cost of concentrating the motion away from that endpoint.
Table~\ref{tab:profiles} lists all six and Figure~\ref{fig:profiles} plots them.

The two sub-families differ in how they distribute the motion along the trajectory.
The \emph{braking} sub-family has monotonically decreasing velocity:
the sample moves fastest near $t=0$ and decelerates to rest at $t=1$.
The \emph{symmetric} sub-family is symmetric about $t=1/2$ with zero velocity at both endpoints.

\begin{table}
\centering
\caption{The six polynomial velocity profiles, each satisfying $\alpha(0)=0$, $\alpha(1)=1$, and $\int_0^1 v(t)\,dt=1$.}
\centering
\label{tab:profiles}
\begin{tabular}{lllcccl}
\toprule
Profile & $\alpha(t)$ & $v(t) = \alpha'(t)$ & $v(0)$ & $v(1)$
        & $v_{\max}$ & Smoothness \\
\midrule
\multicolumn{7}{l}{\textit{Baseline}} \\
Linear
  & $t$
  & $1$
  & $1$ & $1$ & $1$ & $C^\infty$ \\
\midrule
\multicolumn{7}{l}{\textit{Braking (monotonically decreasing velocity)}} \\
Uniform decel.
  & $2t - t^2$
  & $2 - 2t$
  & $2$ & $0$ & $2$ & $C^1$ at $t{=}1$ \\
Coasting
  & $(3t - t^3)/2$
  & $\tfrac{3}{2}(1-t^2)$
  & $3/2$ & $0$ & $3/2$ & $C^1$ at $t{=}1$ \\
Smooth landing
  & $3t - 3t^2 + t^3$
  & $3(1-t)^2$
  & $3$ & $0$ & $3$ & $C^2$ at $t{=}1$ \\
\midrule
\multicolumn{7}{l}{\textit{Symmetric}} \\
Ease-in-out
  & $3t^2 - 2t^3$
  & $6t(1-t)$
  & $0$ & $0$ & $3/2$ & $C^1$ both ends \\
Minimum-jerk
  & $10t^3 - 15t^4 + 6t^5$
  & $30t^2(1-t)^2$
  & $0$ & $0$ & $1.875$ & $C^2$ both ends \\
\bottomrule
\end{tabular}
\end{table}

\paragraph{Linear}
$v(t)=1$, $\alpha(t)=t$. Recovers the standard flow-matching interpolant.

\paragraph{Uniform deceleration}
$\alpha(t)=2t-t^2=1-(1-t)^2$ is the unique quadratic satisfying
$\alpha(0)=0$, $\alpha(1)=1$, and $\alpha'(1)=0$. 
The velocity $v(t)=2-2t$ falls at constant rate $\alpha''=-2$ 
and reaches zero at $t=1$, with $3/4$ of the displacement traversed by $t=1/2$. 
The inverse $\alpha^{-1}(\tau)=1-\sqrt{1-\tau}$ is available in closed form.

\paragraph{Coasting}
$\alpha(t)=(3t-t^3)/2$, $v(t)=\tfrac{3}{2}(1-t^2)$. 
Imposing $\alpha''(0)=0$ in addition to the endpoint conditions yields a gentle
initial deceleration that grows toward $t=1$. 
Peak velocity is $3/2$ and $\alpha(1/2)\approx 0.69$.

\paragraph{Smooth landing}
$\alpha(t)=3t-3t^2+t^3=1-(1-t)^3$, $v(t)=3(1-t)^2$. 
This is the unique cubic with $\alpha'(1)=\alpha''(1)=0$, 
so both velocity and acceleration vanish at $t=1$. 
Its starting velocity $v(0)=3$ is the highest of the six profiles, and $\alpha(1/2)=7/8$.

\paragraph{Ease-in-out}
$\alpha(t)=3t^2-2t^3$, $v(t)=6t(1-t)$. 
This is the cubic Hermite smoothstep, the unique cubic with $\alpha'(0)=\alpha'(1)=0$. 
The bell-shaped velocity peaks at $v_{\max}=3/2$ at $t=1/2$, and by symmetry $\alpha(1/2)=1/2$.

\paragraph{Minimum-jerk}
$\alpha(t)=10t^3-15t^4+6t^5$, $v(t)=30t^2(1-t)^2$. 
This is the unique quintic with $\alpha'(0)=\alpha'(1)=\alpha''(0)=\alpha''(1)=0$. 
Flash and Hogan~\cite{flash1985coordination} showed that the same polynomial
minimizes $\int_0^1|\alpha'''(t)|^2\,dt$ under those boundary conditions;
it is $C^2$ at both endpoints and peaks at $v_{\max}=1.875$.

The braking sub-family tests whether deceleration toward $t=1$ is beneficial as a class, 
and the symmetric sub-family acts as a control: 
similar smoothness, but no directional bias toward the data endpoint.
Section~\ref{sec:method-error} gives a quantitative account of why this direction matters for noise-to-data generation.

\subsection{Inference-time rescheduling}
\label{sec:method-inference}

Any flow-matching model trained under the linear interpolant can be sampled under any admissible velocity profile, 
with no retraining. The construction is given below.

\paragraph{Setup}
Let $\vtheta$ be a network trained under the linear interpolant. 
On its training distribution,
\begin{equation}
  \vtheta(x_\tau, \tau) \approx x_1 - x_0,
  x_\tau = x_0 + \tau\,(x_1-x_0),  \tau \in [0,1].
  \label{eq:pretrained-property}
\end{equation}
Under a velocity profile $v$, the sample at time $t$ sits at $x_t^v = x_0 + \alpha(t)(x_1-x_0)$, 
which does not lie on the linear trajectory at time $t$. 
Querying the network as $\vtheta(x,t)$ would therefore place it out of distribution.

\paragraph{The adapted velocity field}
Define
\begin{equation}
  \vtilde(x,t) = v(t)\, \vtheta(x,\, \alpha(t))
  \label{eq:adapted-field}
\end{equation}
and integrate $dx/dt = \vtilde(x,t)$. 
The linear trajectory $x_\tau^{\mathrm{lin}} = x_0 + \tau(x_1-x_0)$ 
and the VSFM trajectory $x_t^v = x_0 + \alpha(t)(x_1-x_0)$ traverse the same straight-line
segment and differ only in speed. 
Equating $x_t^v = x_\tau^{\mathrm{lin}}$ gives $\tau = \alpha(t)$: 
the spatial point visited at VSFM time $t$ coincides with the one that the linear
trajectory visits at time $\alpha(t)$. 
Substituting into Equation~\eqref{eq:pretrained-property}, $\vtheta(x, \alpha(t)) \approx x_1 - x_0$, 
because the spatial input and time argument now lie jointly on the network's training distribution.
Multiplication by $v(t)$ recovers the target conditional velocity
$v(t)(x_1-x_0)$ of Equation~\eqref{eq:vsfm-conditional}. 
The adapted field therefore matches the desired VSFM velocity up to the residual error of the pretrained network.

\paragraph{Discrete-time form}
The change of variables $\tau = \alpha(t)$, with $d\tau = v(t)\,dt$,
turns the adapted ODE $dx/dt = v(t)\,\vtheta(x,\alpha(t))$ into
\begin{equation}
  \frac{dx}{d\tau} = \vtheta(x,\tau).
  \label{eq:tau-ode}
\end{equation}
This is the pretrained model's own ODE, parameterized by $\tau$.

\paragraph{Displacement-predictor view}
Equation~\eqref{eq:tau-ode} exposes $\vtheta(x,\tau)$ as a \emph{displacement predictor}: 
it outputs the full $x_1-x_0$ rather than an instantaneous rate. 
Both the adaptation procedure and the training loss in Equation~\eqref{eq:vsfm-loss} follow from this reading. 
At inference, the predictor is queried at $\tau=\alpha(t)$ and advanced by $\Delta\tau_i$, 
applying the correct fraction of the predicted displacement. 
At training, the predictor is regressed directly on $x_1-x_0$, 
so the profile enters only through positions during training and through step sizes at inference.

\paragraph{Exact coordinate stepping}
Because $\alpha$ is available in closed form for every polynomial profile we consider, 
the fractional displacement accumulated from $t_i$ to $t_{i+1}$ admits the closed-form expression
\begin{equation}
  \Delta\tau_i = \alpha(t_{i+1}) - \alpha(t_i) = \int_{t_i}^{t_{i+1}} v(t)\,dt,
  \label{eq:dtau}
\end{equation}
with no quadrature approximation. 
Since $\vtheta$ predicts the full displacement $x_1-x_0$, 
multiplying by $\Delta\tau_i$ yields the coordinate increment for that interval. 
The full inference update is
\begin{equation}
  x_{i+1} = x_i + \vtheta(x_i,\,\tau_i)\,\Delta\tau_i,
  \tau_i = \alpha\!\left(\tfrac{i}{N}\right),
  \Delta\tau_i = \alpha\!\left(\tfrac{i+1}{N}\right) - \alpha\!\left(\tfrac{i}{N}\right).
  \label{eq:vsfm-euler}
\end{equation}

\begin{algorithm}[t]
\caption{VSFM inference (exact $\tau$-space Euler).}
\label{alg:vsfm-inference}
\begin{algorithmic}[1]
\Require Pretrained network $\vtheta$; movement profile $\alpha$; steps $N$
\State Sample $x_0 \sim \mathcal{N}(0, I)$
\For{$i = 0, 1, \ldots, N-1$}
    \State $\tau_i \gets \alpha(i/N)$; \quad
           $\Delta\tau_i \gets \alpha((i{+}1)/N) - \alpha(i/N)$
    \State $x_{i+1} \gets x_i + \vtheta(x_i,\, \tau_i)\cdot\Delta\tau_i$
\EndFor
\State \Return $x_N$
\end{algorithmic}
\end{algorithm}

\paragraph{Connection to non-uniform time scheduling}
Equation~\eqref{eq:vsfm-euler} integrates the pretrained model's $\tau$-space ODE 
on the non-uniform grid $\tau_i = \alpha(i/N)$, 
which operationally coincides with running a standard linear-FM Euler sampler on that grid. 
Section~\ref{sec:related-work} discusses how VSFM sits in the broader non-uniform-scheduling literature~\cite{karras2022elucidating,esser2024scaling}.

\paragraph{Effective step sizes}
The per-step displacement $\Delta\tau_i \approx v(t_i)/N$ varies along the trajectory. 
Table~\ref{tab:effective-step-sizes} reports the allocation for each profile. 
Braking profiles take large $\tau$-steps near $t=0$, 
where the field is comparatively smooth, and small $\tau$-steps near $t=1$, 
where the field varies most rapidly; 
the symmetric profiles concentrate their largest steps at the midpoint instead. 
Section~\ref{sec:method-error} makes the intuition behind these choices quantitative.

\begin{table}[t]
\centering
\caption{Approximate $\tau$-step size $\Delta\tau_i \approx v(t_i)/N$ (in units of $1/N$) under each profile on a uniform time grid.}
\label{tab:effective-step-sizes}
\begin{tabular}{lccc}
\toprule
Profile & near $t=0$ & near $t=1/2$ & near $t=1$ \\
\midrule
Linear         & $\approx 1$   & $\approx 1$     & $\approx 1$ \\
Uniform decel. & $\approx 2$   & $\approx 1$     & $\approx 0$ \\
Coasting       & $\approx 1.5$ & $\approx 1.125$ & $\approx 0$ \\
Smooth landing & $\approx 3$   & $\approx 0.75$  & $\approx 0$ \\
Ease-in-out    & $\approx 0$   & $\approx 1.5$   & $\approx 0$ \\
Minimum-jerk   & $\approx 0$   & $\approx 1.875$ & $\approx 0$ \\
\bottomrule
\end{tabular}
\end{table}

\subsection{Discretization error analysis}
\label{sec:method-error}

We now examine how the velocity profile shapes the local truncation error
of the Euler integrator applied to the $\tau$-space ODE $dx/d\tau = \vtheta(x,\tau)$~\cite{chen2018neural}
with non-uniform steps $\Delta\tau_i = \alpha(t_{i+1}) - \alpha(t_i)$ (Equation~\eqref{eq:vsfm-euler}).
All six profiles are treated below, beginning with the linear baseline and uniform deceleration.

For an ODE $dx/d\tau = g(x,\tau)$ integrated by $x_{i+1} = x_i + g(x_i,\tau_i)\,\Delta\tau_i$, 
the local truncation error at step $i$ is
\begin{equation}
  e_i = \frac{(\Delta\tau_i)^2}{2}\,
        \left.\frac{d^2 x}{d\tau^2}\right|_{\tau=\tau_i}
        + \mathcal{O}((\Delta\tau_i)^3).
  \label{eq:euler-lte}
\end{equation}
With $g(x,\tau) = \vtheta(x,\tau)$, the second derivative reduces to
\begin{equation}
  \frac{d^2 x}{d\tau^2}
  = \underbrace{\frac{\partial \vtheta}{\partial x}\,\vtheta
    + \frac{\partial \vtheta}{\partial \tau}}_{\text{field variation}}.
  \label{eq:second-derivative}
\end{equation}
No profile-curvature term appears, because the $\tau$-space ODE carries no explicit $v(t)$ factor; 
the only source of truncation error is the nonlinearity of the learned field $\vtheta$. 
On a uniform $t$-grid the step size satisfies $\Delta\tau_i \approx v(t_i)\,\Delta t$, 
hence $(\Delta\tau_i)^2 \approx v(t_i)^2\,(\Delta t)^2$, 
and the local error scales as $v(t_i)^2$ times the field variation at $\tau_i$.

\paragraph{Linear profile}
Under $v(t)\equiv 1$, $\Delta\tau_i = \Delta t$ uniformly and $(\Delta\tau_i)^2 = (\Delta t)^2$. 
The field-variation term in Equation~\eqref{eq:second-derivative} is not damped anywhere on $[0,1]$. 
Near $\tau=1$, where learned flow-matching fields tend to vary rapidly as the trajectory commits 
to a particular data sample, the truncation error is at full magnitude.

\paragraph{Uniform deceleration.}
Under $v(t)=2-2t$, the step $\Delta\tau_i \approx 2(1-t_i)\,\Delta t$ shrinks to zero as $t\to 1$ ($\tau\to 1$), 
so $(\Delta\tau_i)^2 \approx 4(1-t_i)^2(\Delta t)^2 \to 0$. 
Field-variation error is therefore damped where the learned field varies most. 
The absence of a profile-curvature term in the $\tau$-space formulation means that no compensating error term appears.

\paragraph{Coasting}
Under $v(t)=\tfrac{3}{2}(1-t^2)$, the start $v(0)=3/2$ gives $(\Delta\tau_i)^2\approx(9/4)(\Delta t)^2$,
milder than uniform deceleration's $4(\Delta t)^2$.
Near $t=1$, the factorization $(1-t^2)=(1-t)(1+t)\to 2(1-t)$ gives $(\Delta\tau_i)^2\approx 9(1-t_i)^2(\Delta t)^2\to 0$:
the same damping at the data manifold as uniform deceleration, but with a smaller starting step.
This matches the experimental finding that coasting ranks above uniform deceleration at every NFE budget.

\paragraph{Smooth landing}
Under $v(t)=3(1-t)^2$, the start $v(0)=3$ gives $(\Delta\tau_i)^2\approx 9(\Delta t)^2$,
the largest initial squared step of the six profiles.
Near $t=1$, $(\Delta\tau_i)^2\approx 9(1-t_i)^4(\Delta t)^2\to 0$, the fastest suppression of any braking profile.
The near-data precision is bought with a large initial step that injects substantial truncation error early,
the mechanism behind the failure mode noted in Section~\ref{sec:exp-analysis}.

\paragraph{Ease-in-out}
Under $v(t)=6t(1-t)$, the error is damped at both endpoints ($v\to 0$ linearly as $t\to 0$ and $t\to 1$),
while the midpoint $v(1/2)=3/2$ gives $(\Delta\tau_i)^2\approx(9/4)(\Delta t)^2$.
The largest steps land at the midpoint, where the field's nonlinearity is lower than near $\tau=1$,
so for noise-to-data generation the profile yields a uniform but small improvement over the linear baseline.

\paragraph{Minimum-jerk}
Under $v(t)=30t^2(1-t)^2$, the velocity vanishes quadratically at both endpoints, a faster decay than ease-in-out,
while the midpoint $v(1/2)\approx 1.875$ gives the largest midpoint step of the six profiles, $(\Delta\tau_i)^2\approx 3.52(\Delta t)^2$.
The picture follows ease-in-out, only more pronounced at every location,
and empirically the two behave similarly, with a slight minimum-jerk deficit at very low NFE where the midpoint step dominates the error budget.

\paragraph{Summary.}
A velocity profile that places small $v(t_i)^2$, and therefore small $(\Delta\tau_i)^2$, 
where the learned field $\vtheta$ varies most rapidly lowers the local Euler truncation error 
relative to the linear baseline. 
Among the six profiles, the braking sub-family has this structure with respect to $\tau=1$ 
and the symmetric sub-family does not.
The corresponding prediction is that braking profiles improve over the baseline,
with the largest gain at low NFE, where truncation error dominates,
and shrinking gains at high NFE, where it does not.
The experiments of Section~\ref{sec:experiments} confirm this pattern.
The picture changes when the source $x_0$ is itself structured, as in image-to-image tasks:
field variation near $\tau=0$ becomes significant,
and the bilateral suppression of the symmetric profiles becomes useful.

\section{Experiments}
\label{sec:experiments}

\subsection{Experimental setup}
\label{sec:exp-setup}

Every experiment uses CIFAR-10 ($32{\times}32$, 50{,}000 training images with random horizontal flips). 
Sampling is carried out with the fixed-step Euler integrator of Algorithm~\ref{alg:vsfm-inference} 
on a uniform time grid. 
Fr\'echet Inception Distance~(FID) \cite{heusel2017gans} is reported against $10{,}000$ held-out validation images.

\paragraph{Training from scratch}
For each of the six velocity profiles we train a class-conditional UNet adapted from OpenAI's guided-diffusion code~\cite{dhariwal2021diffusion}
and configured for $32\times 32$ images in the same lineage as TorchCFM~\cite{tong2024improving}: 
base channel width $128$, channel multipliers $(1, 2, 2, 2)$, two residual blocks per resolution, 
and a self-attention block at the lowest spatial resolution. 
The model is conditioned on the 10 CIFAR-10 class labels. 
The six profiles share an identical optimization recipe: 
Adam with learning rate $2\times 10^{-4}$, batch size $128$, 
an exponential moving average of weights with decay $0.9999$, 
and a cosine learning-rate schedule. 
Each profile is trained for $187{,}500$ gradient steps with independent couplings $(x_0, x_1)$.

\paragraph{Inference-time adaptation}
Each velocity profile is applied to the publicly available pretrained TorchCFM CIFAR-10 checkpoint~\cite{tong2024improving} 
via Algorithm~\ref{alg:vsfm-inference}, with no retraining. 
The checkpoint was trained under the linear interpolant. 
The FID values reported by Tong et al.\ for this checkpoint use an adaptive Dormand-Prince (dopri5) solver, 
whereas we evaluate \emph{all} profiles, including the linear baseline, with the same fixed-step Euler integrator. 
Our linear-baseline numbers therefore differ from theirs, but the comparison in Section~\ref{sec:exp-inference} 
isolates the effect of the velocity profile under a common solver. 

\subsection{Training with velocity profiles}
\label{sec:exp-training}

Table~\ref{tab:training-fid} reports the end-of-training FID for the six profiles.

\begin{table}[t]
\caption{Training-from-scratch FID on CIFAR-10. Lower is better. Best result per column in \textbf{bold}.}
\label{tab:training-fid}
\centering
\begin{tabular}{lccc}
\toprule
Profile & 4 NFE & 8 NFE & 50 NFE \\
\midrule
\multicolumn{4}{l}{\textit{Baseline}} \\
Linear & 42.86 & 20.36 & 9.25 \\
\midrule
\multicolumn{4}{l}{\textit{Braking}} \\
Uniform deceleration & 37.13 & 17.74 & 8.44 \\
Coasting              & \textbf{35.39} & \textbf{17.12} & \textbf{8.29} \\
Smooth landing        & 53.02 & 22.38 & 9.36 \\
\midrule
\multicolumn{4}{l}{\textit{Symmetric}} \\
Ease-in-out    & 39.73 & 18.76 & 8.62 \\
Minimum-jerk   & 45.34 & 21.62 & 9.02 \\
\bottomrule
\end{tabular}
\end{table}

Coasting attains the best FID at every NFE budget: $35.39$ at 4 NFE ($-17.4\%$ over linear), 
$17.12$ at 8 NFE ($-15.9\%$), and $8.29$ at 50 NFE ($-10.4\%$). 
Uniform deceleration is the second-best braking profile, 
and both improve on the linear baseline at every budget. 
The margin is largest at low NFE and contracts as NFE grows, 
which is in line with the discretization-error analysis of Section~\ref{sec:method-error}: 
at large NFE the step sizes are small enough that the profile choice has limited bite.

Smooth landing is the only braking profile that underperforms, 
trailing linear at all three budgets ($53.02$ vs.\ $42.86$ at 4 NFE; $9.36$ vs.\ $9.25$ at 50 NFE). 
Its inference-time behavior follows the same pattern; 
Section~\ref{sec:exp-analysis} examines the mechanism.

The symmetric profiles (ease-in-out, minimum-jerk) bring modest improvements over linear at 50 NFE 
but trail it at 4 NFE. 
Their bell-shaped velocity concentrates integration effort at the trajectory midpoint rather than near the data endpoint.

\begin{figure}[ht]
\centering
\includegraphics{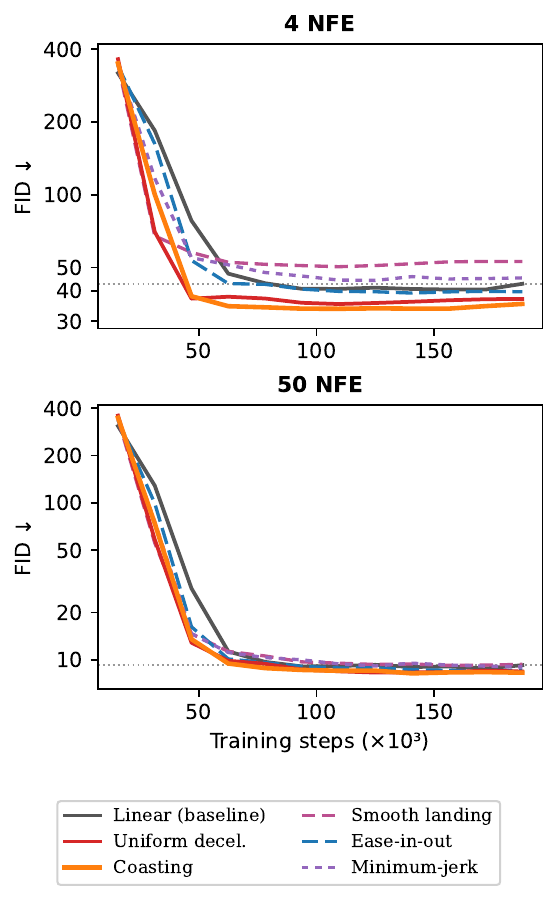}
\caption{FID vs.\ training steps on CIFAR-10 (training from scratch).
Top: 4 NFE; bottom: 50 NFE; log scale. The dotted line marks the linear baseline's final FID.}
\label{fig:convergence}
\end{figure}

Figure~\ref{fig:convergence} plots FID against training steps at 4 and 50 NFE. 
In both panels the braking profiles (coasting, uniform deceleration) 
descend faster than linear during early training. 
At 4 NFE, coasting falls below the linear baseline's final FID of $42.86$ by step $\approx47{,}000$, 
roughly a quarter of the total training budget, while linear reaches this quality only at the end of training. 
At 50 NFE all profiles converge to similar final values, 
but coasting and uniform deceleration reach linear's final FID of $\approx9.25$
some $15{,}000$ steps earlier ($\approx78{,}000$ vs.\ $\approx93{,}000$ steps).

\subsection{Inference-time adaptation}
\label{sec:exp-inference}

\begin{figure}[ht]
\centering
\includegraphics{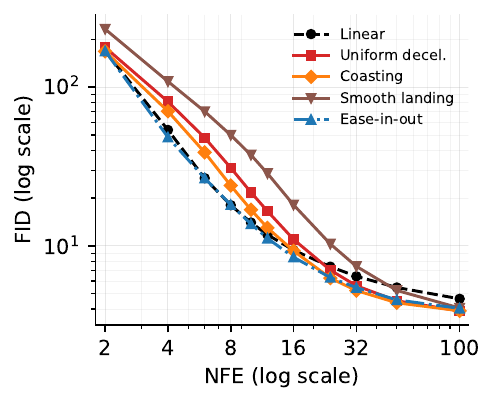}
\caption{Inference-time FID vs.\ NFE on CIFAR-10 (pretrained TorchCFM~\cite{tong2024improving},
Euler integration of Algorithm~\ref{alg:vsfm-inference}, no retraining),
for the linear baseline, the three braking profiles, and the best-performing symmetric profile (ease-in-out).}
\label{fig:inference-curves}
\end{figure}

Table~\ref{tab:inference-fid} reports FID at 32, 50, and 100 NFE 
for the six profiles applied to the pretrained TorchCFM checkpoint~\cite{tong2024improving}.
Figure~\ref{fig:inference-curves} plots the FID-vs-NFE curves across the 11 budgets we evaluated, 
restricted to the linear baseline, the three braking profiles, and the best-performing symmetric profile.

\begin{table}[ht]
\caption{Inference-time FID on CIFAR-10 (pretrained TorchCFM, no retraining).
Lower is better. Best result per column in \textbf{bold}.}
\label{tab:inference-fid}
\centering
\begin{tabular}{lccc}
\toprule
Profile & 32 NFE & 50 NFE & 100 NFE \\
\midrule
\multicolumn{4}{l}{\textit{Baseline}} \\
Linear & 6.43 & 5.47 & 4.64 \\
\midrule
\multicolumn{4}{l}{\textit{Braking}} \\
Uniform deceleration & 5.61 & 4.51 & \textbf{3.89} \\
Coasting              & \textbf{5.23} & \textbf{4.38} & 3.90 \\
Smooth landing        & 7.43 & 5.24 & 4.06 \\
\midrule
\multicolumn{4}{l}{\textit{Symmetric}} \\
Ease-in-out    & 5.46 & 4.57 & 4.04 \\
Minimum-jerk   & 5.64 & 4.61 & 4.12 \\
\bottomrule
\end{tabular}
\end{table}

All braking profiles except smooth landing outperform the linear baseline at every NFE budget. 
Coasting attains the best FID at 32 and 50 NFE: $5.23$ ($-18.6\%$ over linear) and $4.38$ ($-19.8\%$),
while uniform deceleration leads at 100 NFE with $3.89$ ($-16.0\%$).
The symmetric profiles deliver smaller but still uniformly positive improvements.

Smooth landing is the exception at the lowest budget: 
its FID at 32 steps, $7.43$, exceeds the linear baseline's $6.43$.
The step-size allocation in Table~\ref{tab:effective-step-sizes} explains the discrepancy. 
Smooth landing has $v(0)=3$, the largest starting velocity of the six profiles, 
so the earliest Euler steps cover three times the $\tau$-interval of the linear baseline. 
At 32 NFE those large initial steps introduce truncation error that outweighs 
the benefit of the fine-grained final steps near $t=1$. 
As the step budget grows, the per-step error shrinks and the benefit of small $\Delta\tau_i$ 
near the data manifold takes over: 
smooth landing reaches $5.24$ at 50 NFE and $4.06$ at 100 NFE, a $12.3\%$ improvement over linear.

Taken together, the results support the training-free adaptation mechanism of Section~\ref{sec:method-inference}. 
A network trained entirely under the linear interpolant, 
queried at remapped times $\tau_i = \alpha(t_i)$ 
and stepped by exact increments $\Delta\tau_i = \alpha(t_{i+1})-\alpha(t_i)$, 
produces uniformly lower FID for braking profiles at moderate to large NFE budgets, 
at no training cost.

\subsection{Analysis}
\label{sec:exp-analysis}

\paragraph{Which profiles help and why}
The braking sub-family exhibits the same pattern in both experimental settings: 
profiles with monotonically decreasing velocity improve on linear, 
the sole exception being smooth landing (treated below), 
while the symmetric profiles deliver only marginal gains. 
This is the prediction of the step-size analysis in Table~\ref{tab:effective-step-sizes}. 
Braking profiles assign small $\Delta\tau_i$ near $t=1$, the data manifold, 
where the velocity field $\vtheta(x,\tau)$ is most nonlinear, 
and so suppress the dominant contribution to Euler truncation error.

\paragraph{The smooth-landing exception}
Smooth landing trails linear in both training (every NFE) and inference (32 NFE). 
Its high initial velocity $v(0)=3$ allocates the largest $\tau$-steps to the near-noise portion of the trajectory, 
where the field is relatively smooth, 
and leaves only a handful of small steps for $t\approx 1$, where precision matters most. 
At 32 NFE the budget spent on the initial steps is wasted; 
the profile becomes beneficial only once the step count is large enough that those final small steps dominate the error.
Coasting, with the milder $v(0)=3/2$, decelerates more gently and strikes a better balance, 
which is why it ranks first across all settings tested.

\section{Conclusions}
\label{sec:conclusions}

This paper introduced Velocity Scheduled Flow Matching (VSFM), 
a generalization of flow matching in which 
the constant conditional velocity target $x_1 - x_0$ 
is replaced by the time-varying form $v(t)(x_1-x_0)$ 
for any nonnegative profile $v$ with $\int_0^1 v\,dt=1$.
The framework supports two distinct uses. 
At inference time, integrating the pretrained model's $\tau$-space ODE (Equation~\eqref{eq:tau-ode}) 
with exact steps $\Delta\tau_i = \alpha(t_{i+1})-\alpha(t_i)$ applies any admissible velocity profile 
to an existing linear flow-matching model with no retraining; 
the coasting profile lowers FID by up to $19.8\%$ on CIFAR-10. 
At training time, fitting the network under a non-constant profile delivers a further drop in FID 
of $17.4\%$ at 4~NFE and $10.4\%$ at 50~NFE relative to the linear baseline. 
A single mechanism underlies both effects: 
braking profiles damp the field-variation term 
$(\Delta\tau_i)^2 \cdot \partial^2 x/\partial\tau^2$ of the Euler local truncation error near $\tau=1$, 
the region where the learned velocity field varies most rapidly.

Taken together, these results identify the velocity profile as a simple and low-cost lever
for improving few-step sampling in flow matching,
and the present study, conducted on CIFAR-10 ($32{\times}32$) with six polynomial profiles under a fixed-step Euler sampler,
opens several directions for further work.
Replication on higher-resolution and more diverse image datasets would establish how broadly the observed gains extend.
The discretization-error analysis developed here is a local argument;
a global convergence bound relating the properties of a profile to the total integration error
would place the mechanism on a firmer theoretical footing.
The profile family can be broadened beyond the polynomial case to trigonometric and sigmoidal forms,
or learned jointly with the network weights instead of being fixed in advance.
The interaction between velocity scheduling and higher-order solvers such as DPM-Solver~\cite{lu2022dpm} and DDIM~\cite{song2020denoising},
which cancel the leading truncation error, also merits study,
as it would clarify the regimes in which the choice of profile contributes most.
VSFM schedules could further be combined with progressive distillation~\cite{salimans2022progressive}
or consistency training to reduce sampling cost.
Finally, velocity profiles may transfer to scientific generation tasks
such as molecular dynamics~\cite{klein2023equivariant} and protein structure prediction~\cite{bose2024foldflow},
where the structure of the data manifold differs from that of natural images.


\bibliographystyle{unsrt}  
\bibliography{references}

@article{lipman2022flow,
  title={Flow matching for generative modeling},
  author={Lipman, Yaron and Chen, Ricky TQ and Ben-Hamu, Heli and Nickel, Maximilian and Le, Matt},
  journal={arXiv preprint arXiv:2210.02747},
  year={2022},
  langid = {english},
}

@article{liu2022flow,
  title={Flow straight and fast: Learning to generate and transfer data with rectified flow},
  author={Liu, Xingchao and Gong, Chengyue and Liu, Qiang},
  journal={arXiv preprint arXiv:2209.03003},
  year={2022},
  langid = {english},
}

@article{tong2024improving,
title={Improving and generalizing flow-based generative models with minibatch optimal transport},
author={Alexander Tong and Kilian FATRAS and Nikolay Malkin and Guillaume Huguet and Yanlei Zhang and Jarrid Rector-Brooks and Guy Wolf and Yoshua Bengio},
journal={Transactions on Machine Learning Research},
issn={2835-8856},
year={2024},
langid = {english},
}

@inproceedings{shaul2023kinetic,
  title={On kinetic optimal probability paths for generative models},
  author={Shaul, Neta and Chen, Ricky TQ and Nickel, Maximilian and Le, Matthew and Lipman, Yaron},
  booktitle={International Conference on Machine Learning},
  pages={30883-30907},
  year={2023},
  organization={PMLR}
}

@inproceedings{esser2024scaling,
  title={Scaling rectified flow transformers for high-resolution image synthesis},
  author={Esser, Patrick and Kulal, Sumith and Blattmann, Andreas and Entezari, Rahim and M{\"u}ller, Jonas and Saini, Harry and Levi, Yam and Lorenz, Dominik and Sauer, Axel and Boesel, Frederic and others},
  booktitle={Forty-first international conference on machine learning},
  year={2024},
  langid = {english},
}

@article{ho2020denoising,
  title={Denoising diffusion probabilistic models},
  author={Ho, Jonathan and Jain, Ajay and Abbeel, Pieter},
  journal={Advances in neural information processing systems},
  volume={33},
  pages={6840-6851},
  year={2020},
  langid = {english},
}

@inproceedings{nichol2021improved,
  title={Improved denoising diffusion probabilistic models},
  author={Nichol, Alexander Quinn and Dhariwal, Prafulla},
  booktitle={International conference on machine learning},
  pages={8162-8171},
  year={2021},
  organization={PMLR},
  langid = {english},
}

@article{karras2022elucidating,
  title={Elucidating the design space of diffusion-based generative models},
  author={Karras, Tero and Aittala, Miika and Aila, Timo and Laine, Samuli},
  journal={Advances in neural information processing systems},
  volume={35},
  pages={26565-26577},
  year={2022},
  langid = {english},
}

@inproceedings{choi2022perception,
  title={Perception prioritized training of diffusion models},
  author={Choi, Jooyoung and Lee, Jungbeom and Shin, Chaehun and Kim, Sungwon and Kim, Hyunwoo and Yoon, Sungroh},
  booktitle={Proceedings of the IEEE/CVF conference on computer vision and pattern recognition},
  pages={11472-11481},
  year={2022}
}

@inproceedings{hang2023efficient,
  title={Efficient diffusion training via min-snr weighting strategy},
  author={Hang, Tiankai and Gu, Shuyang and Li, Chen and Bao, Jianmin and Chen, Dong and Hu, Han and Geng, Xin and Guo, Baining},
  booktitle={Proceedings of the IEEE/CVF international conference on computer vision},
  pages={7441-7451},
  year={2023}
}

@article{kingma2023understanding,
  title={Understanding diffusion objectives as the elbo with simple data augmentation},
  author={Kingma, Diederik and Gao, Ruiqi},
  journal={Advances in Neural Information Processing Systems},
  volume={36},
  pages={65484-65516},
  year={2023}
}

@article {flash1985coordination,
	author = {Flash, T and Hogan, N},
	title = {The coordination of arm movements: an experimentally confirmed mathematical model},
	volume = {5},
	number = {7},
	pages = {1688-1703},
	year = {1985},
	doi = {10.1523/JNEUROSCI.05-07-01688.1985},
	publisher = {Society for Neuroscience},
	issn = {0270-6474},
	URL = {https://www.jneurosci.org/content/5/7/1688},
	eprint = {https://www.jneurosci.org/content/5/7/1688.full.pdf},
	journal = {Journal of Neuroscience}
}

@inproceedings{bondar2025deep,
  title={Deep generative models as the probability transformation functions},
  author={Bondar, Vitalii and Babenko, Vira and Trembovetskyi, Roman and Korobeinyk, Yurii and Dzyuba, Viktoriya},
  booktitle={International Conference on Information and Software Technologies},
  pages={97-111},
  year={2025},
  organization={Springer}
}

@article{bondar2026inference,
  title={Inference-time Scaling as a Universal Principle in Machine Learning: Classification and Cross-domain Analysis},
  author={Bondar, Vitalii},
  journal={Visnyk of Kherson National Technical University},
  number={1 (96)},
  pages={200-206},
  year={2026},
  langid = {english},
}

@book{craig2009introduction,
  title={Introduction to robotics: mechanics and control, 3/E},
  author={Craig, John J},
  year={2009},
  publisher={Pearson Education India}
}

@article{yang2024consistencyfm,
  title={Consistency flow matching: Defining straight flows with velocity consistency},
  author={Yang, Ling and Zhang, Zixiang and Zhang, Zhilong and Liu, Xingchao and Xu, Minkai and Zhang, Wentao and Meng, Chenlin and Ermon, Stefano and Cui, Bin},
  journal={arXiv preprint arXiv:2407.02398},
  year={2024}
}

@article{rathod2026splineflow,
  title={SplineFlow: Flow Matching for Dynamical Systems with B-Spline Interpolants},
  author={Rathod, Santanu Subhash and Li{\`o}, Pietro and Zhang, Xiao},
  journal={arXiv preprint arXiv:2601.23072},
  year={2026}
}

@article{chen2025densejump,
  title={Dense-Jump Flow Matching with Non-Uniform Time Scheduling for Robotic Policies: Mitigating Multi-Step Inference Degradation},
  author={Chen, Zidong and Guo, Zihao and Wang, Peng and Egbe, ThankGod Itua and Lyu, Yan and Qian, Chenghao},
  journal={arXiv preprint arXiv:2509.13574},
  year={2025},
  langid={english}
}

@article{heusel2017gans,
  title={Gans trained by a two time-scale update rule converge to a local nash equilibrium},
  author={Heusel, Martin and Ramsauer, Hubert and Unterthiner, Thomas and Nessler, Bernhard and Hochreiter, Sepp},
  journal={Advances in neural information processing systems},
  volume={30},
  year={2017}
}

@InProceedings{pmlr-v37-sohl-dickstein15,
  title = 	 {Deep Unsupervised Learning using Nonequilibrium Thermodynamics},
  author = 	 {Sohl-Dickstein, Jascha and Weiss, Eric and Maheswaranathan, Niru and Ganguli, Surya},
  booktitle = 	 {Proceedings of the 32nd International Conference on Machine Learning},
  pages = 	 {2256-2265},
  year = 	 {2015},
  editor = 	 {Bach, Francis and Blei, David},
  volume = 	 {37},
  series = 	 {Proceedings of Machine Learning Research},
  address = 	 {Lille, France},
  month = 	 {07},
  publisher =    {PMLR},
  langid = {english},
}

@inproceedings{song2021scorebased,
  title={Score-Based Generative Modeling through Stochastic Differential Equations},
  author={Yang Song and Jascha Sohl-Dickstein and Diederik P Kingma and Abhishek Kumar and Stefano Ermon and Ben Poole},
  booktitle={International Conference on Learning Representations},
  year={2021},
  langid = {english},
}

@inproceedings{song2020denoising,
  title={Denoising diffusion implicit models},
  author={Song, Jiaming and Meng, Chenlin and Ermon, Stefano},
  booktitle={International Conference on Learning Representations},
  year={2021},
  langid = {english},
}

@article{albergo2022building,
  title={Building normalizing flows with stochastic interpolants},
  author={Albergo, Michael S and Vanden-Eijnden, Eric},
  journal={arXiv preprint arXiv:2209.15571},
  year={2022}
}

@article{dhariwal2021diffusion,
  title={Diffusion models beat GANs on image synthesis},
  author={Dhariwal, Prafulla and Nichol, Alexander},
  journal={Advances in neural information processing systems},
  volume={34},
  pages={8780-8794},
  year={2021},
  langid = {english},
}

@inproceedings{rombach2022high,
  title={High-resolution image synthesis with latent diffusion models},
  author={Rombach, Robin and Blattmann, Andreas and Lorenz, Dominik and Esser, Patrick and Ommer, Bj{\"o}rn},
  booktitle={Proceedings of the IEEE/CVF conference on computer vision and pattern recognition},
  pages={10684-10695},
  year={2022},
  langid = {english},
}

@article{chen2018neural,
  title={Neural ordinary differential equations},
  author={Chen, Ricky TQ and Rubanova, Yulia and Bettencourt, Jesse and Duvenaud, David K},
  journal={Advances in neural information processing systems},
  volume={31},
  year={2018},
  langid = {english},
}

@inproceedings{lu2022dpm,
  title={{DPM-Solver}: A fast {ODE} solver for diffusion probabilistic model sampling in around 10 steps},
  author={Lu, Cheng and Zhou, Yuhao and Bao, Fan and Chen, Jianfei and Li, Chongxuan and Zhu, Jun},
  booktitle={Advances in Neural Information Processing Systems},
  volume={35},
  pages={5775-5787},
  year={2022},
  langid = {english},
}

@article{salimans2022progressive,
  title={Progressive distillation for fast sampling of diffusion models},
  author={Salimans, Tim and Ho, Jonathan},
  journal={arXiv preprint arXiv:2202.00512},
  year={2022},
  langid = {english},
}

@inproceedings{pooladian2023multisample,
  title={Multisample Flow Matching: Straightening Flows with Minibatch Couplings},
  author={Pooladian, Aram-Alexandre and Ben-Hamu, Heli and Domingo-Enrich, Carles and Amos, Brandon and Lipman, Yaron and Chen, Ricky T Q},
  booktitle={International Conference on Machine Learning},
  pages={28100-28127},
  year={2023},
  organization={PMLR},
  langid={english}
}

@article{klein2023equivariant,
  title={Equivariant flow matching},
  author={Klein, Leon and Kr{\"a}mer, Andreas and No{\'e}, Frank},
  journal={Advances in Neural Information Processing Systems},
  volume={36},
  pages={59886-59910},
  year={2023}
}

@inproceedings{bose2024foldflow,
  title={Se (3)-stochastic flow matching for protein backbone generation},
  author={Bose, Joey and Akhound-Sadegh, Tara and Huguet, Guillaume and Fatras, Kilian and Rector-Brooks, Jarrid and Liu, Chenghao and Nica, Andrei and Korablyov, Maksym and Bronstein, Michael and Tong, Alexander},
  booktitle={International Conference on Learning Representations},
  volume={2024},
  pages={22590-22621},
  year={2024}
}

\end{document}